\documentclass[10pt,conference,a4paper]{IEEEtran}

\usepackage{style}
\usepackage{amsmath,amssymb}
\usepackage{graphicx}
\usepackage{booktabs}
\usepackage{hyperref}
\usepackage{cite}

\title{Learning Urban Access Costs from Origin--Destination Flows via Inverse Optimal Transport}

\author{
  \IEEEauthorblockN{Paula Joy B. Martinez and Sebastian Felipe R. Bundoc}
  \IEEEauthorblockA{Education Center for AI Research, Philippines}
}

\begin{document}

\maketitle

\begin{abstract}
Cities deliver basic services through mixed public-private facility networks, including schools, clinics, transit providers, and subsidized service points. In these systems, planners often observe where households go, but not the latent cost function through which they trade off factors such as distance, price, and institutional access. We study this urban problem through school choice in the Philippines, where the country's largest national education subsidy is intended to redirect learners from congested public schools to participating private schools. Treating school-to-school enrollment flows as an entropic optimal transport plan, we recover latent choice costs using two complementary inverse optimal transport models: an interpretable distance-banded model with a subsidy term, and a neural cost model trained through a differentiable Sinkhorn forward pass. Applied to 283{,}016 learner trips across 23{,}820 observed flows in the most populated region, the framework estimates a subsidy-equivalent distance, $\lambda^{(k)}$, interpreted as the kilometers of perceived travel cost offset by the subsidy. The case demonstrates how administrative origin-destination data can be transformed into interpretable planning metrics for accessibility-aware subsidy design, facility siting, and urban service allocation.
\end{abstract}

\section{Introduction}

Many urban services are delivered through mixed networks of public and private institutions. Education, health care, and social services increasingly depend on subsidies and distributed facilities that ask households to choose among spatially uneven options. In such systems, access is determined not only by nominal price or proximity to the nearest facility, but by a latent cost function through which households trade off distance, affordability, and institutional quality. However, planners typically observe the resulting origin-destination flows but not the behavioral cost surface that produced them.

School choice provides a concrete instance of this problem. In the Philippines, the largest national education subsidy program funds enrollment in participating private junior high schools, partly to relieve congestion in overcrowded public schools. Yet a subsidy can only decongest the public system if the schools it makes affordable are also spatially reachable. Existing approaches model flows with forward gravity or discrete-choice models~\cite{paaral_kdd}, which are useful for prediction but do not directly answer the planning question that matters: \emph{how many kilometers of perceived access does a subsidy buy, and does that value change across spatial regimes?}

We address this gap with inverse optimal transport (OT). Given observed school-to-school flows, we recover the cost function under which those flows are plausible and translate recovered costs into a policy-facing measure of subsidy reach. The paper makes three contributions. First, we introduce two complementary inverse OT estimators: an interpretable distance-banded piecewise model with explicit subsidy terms, and a neural cost model trained through a differentiable Sinkhorn forward pass. Second, we demonstrate the framework on large-scale Philippine school-choice flows as a case study of subsidy-mediated access in a mixed public--private service network. Third, we derive the subsidy-equivalent distance $\lambda^{(k)}$, which measures how many kilometers of perceived travel cost a subsidy offsets in each distance band — a metric directly usable for subsidy calibration, facility placement, and accessibility-aware service allocation.

\section{Method and Data}

\begin{figure*}[t]     
\centering     
\includegraphics[width=\textwidth]{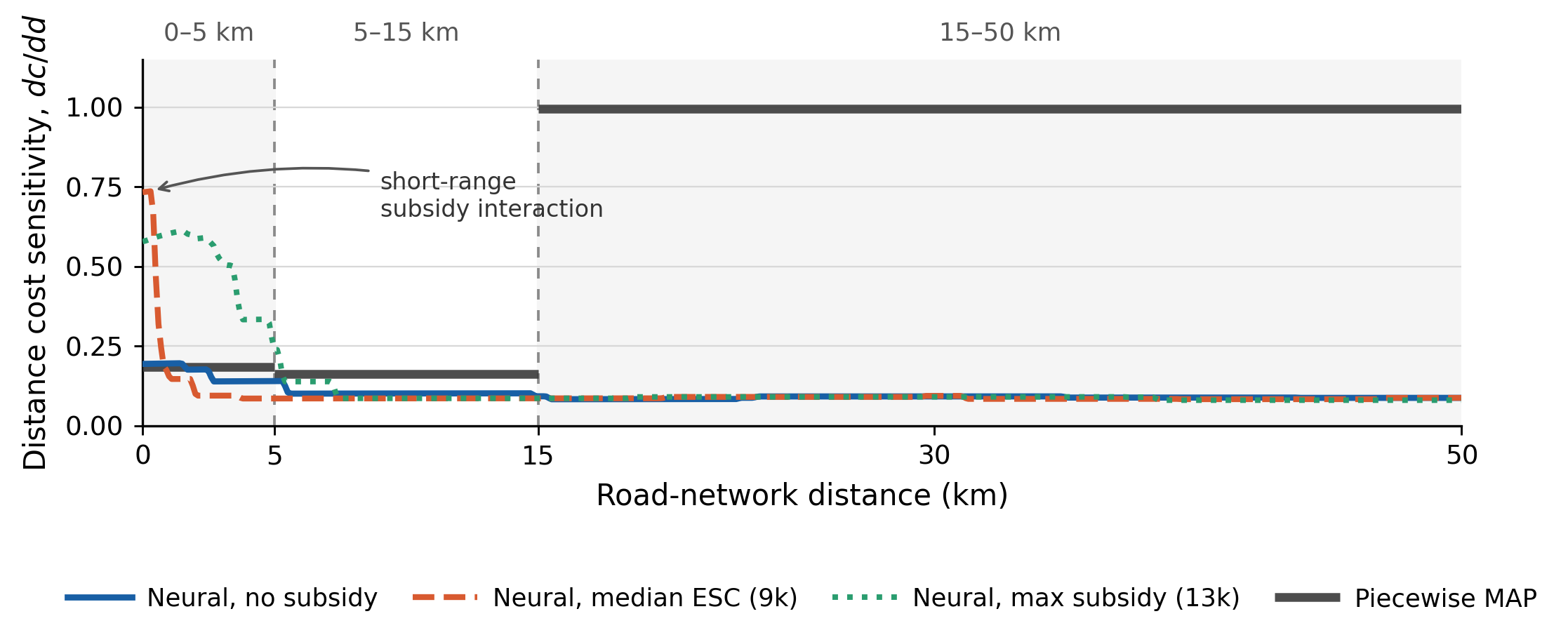}     \caption{Recovered distance-cost sensitivity. Horizontal segments show piecewise MAP estimates; neural curves show $dc/dd$ at zero subsidy, the median positive subsidy of 9{,}000 pesos, and the maximum subsidy of 13{,}000 pesos. The neural model improves fit by 13.7\% and suggests a short-range distance--subsidy interaction rather than a sharp long-range threshold.}     
\label{fig:cost} 
\end{figure*}

Let $T_{\mathrm{obs}}$ be the observed school-to-school learner flow matrix, normalized as a transport plan with origin marginals $\mathbf{p}$ and destination marginals $\mathbf{q}$. Given a cost matrix $C$, the forward model is the entropic OT solution
\[
    \hat{T}=\operatorname{Sinkhorn}(\mathbf{p},\mathbf{q},C,\varepsilon),
    \qquad \varepsilon=0.1.
\]
The inverse problem estimates the cost function that minimizes
\[
    \Phi=\frac{\|T_{\mathrm{obs}}-\hat{T}\|_F^2}{2\sigma^2}.
\]

We estimate two complementary cost models. The first is an interpretable piecewise model,
\[
    c_{ij}=\theta_d^{(k(d_{ij}))}d_{ij}+\theta_e e_j,
\]
where $d_{ij}$ is road-network distance, $e_j$ is the destination subsidy in thousands of pesos, and $k(d_{ij})$ indexes 0--5, 5--15, and 15--50 km bands. The policy quantity is
\[
    \lambda^{(k)}=-\frac{\theta_e}{\theta_d^{(k)}},
\]
interpreted as kilometers of perceived travel cost offset by a 1{,}000-peso subsidy increase. The second model is a neural cost function, $c_{ij}=f_w(d_{ij},e_j)$, estimated by backpropagating through the differentiable Sinkhorn operator.

The case study uses data from the Philippine Department of Education for school years 2022-2024 in the most populous region. We aggregate individual records to school-to-school OD flows and link destinations to school coordinates, road-network distances, and subsidy amounts. The working matrix contains 4{,}200 origin schools, 1{,}861 destination schools, 1{,}852{,}177 feasible OD pairs within 50 km, and 23{,}820 observed positive OD pairs representing 283{,}016 learner trips. All flows are aggregate; no individual learner is identifiable.

\section{Preliminary Results}

The piecewise model was estimated by GPU-accelerated maximum a posteriori (MAP) optimization with a log-domain Sinkhorn forward pass. The likelihood scale was calibrated so that the initial parameter setting had $\Phi=5.00$; the piecewise MAP model reduced this to $\Phi(\theta_{\mathrm{MAP}})=3.29$. The estimated coefficients are $\theta_d^{(\text{1})}=0.185$, $\theta_d^{(\text{2})}=0.163$, $\theta_d^{(\text{3})}=0.994$, and $\theta_e=-0.989$. These imply subsidy-equivalent distances of $\lambda^{(\text{1})}=5.35$ km, $\lambda^{(\text{2})}=6.07$ km, and $\lambda^{(\text{3})}=1.00$ km per 1{,}000 pesos. For example, in the 5--15 km band, a 1{,}000-peso increase in subsidy offsets roughly 6.07 km of perceived travel cost; the subsidy makes a farther school appear closer in the recovered choice-cost function, not that learners physically travel less.

The neural inverse OT model improves fit, achieving $\Phi(w^*)=2.84$, a 13.7\% reduction relative to the piecewise model. Figure~\ref{fig:cost} compares the piecewise coefficients with the neural distance derivative, $dc/dd$, evaluated at no subsidy, the median positive subsidy of 9{,}000 pesos, and the maximum subsidy of 13{,}000 pesos. At zero subsidy, neural distance sensitivity is smooth and low, ranging from 0.083 to 0.196. At the median positive subsidy, it ranges from 0.083 to 0.736; at the maximum subsidy, from 0.080 to 0.612. In both subsidized cases, the largest values occur at very short distances and flatten over medium and longer ranges.

These results show why the two-model design matters. The piecewise model provides an interpretable planning metric, $\lambda^{(k)}$, while the neural model reveals a finer-grained distance--subsidy interaction. In particular, the neural derivative does not reproduce the sharp long-range increase implied by the broad 15--50 km piecewise band, suggesting that some apparent long-distance friction may reflect band aggregation.

\section{Discussion and Impact on Urban Systems}

The central planning implication is that a subsidy has a spatial footprint. Its effect depends not only on the nominal amount, but also on where subsidized facilities are located relative to demand and how distance-sensitive households are in different parts of the network. In the Philippine school-choice case, a uniform subsidy may therefore produce unequal accessibility gains: effective where participating schools are nearby, weaker where subsidized capacity is spatially mismatched with congested public-school catchments.

The framework generalizes beyond education. Health facility choice, transit ridership, subsidized housing search, market access, and evacuation behavior can all be represented as OD matrices with latent spatial frictions. Inverse OT provides a reusable Urban AI approach for learning interpretable behavioral costs from administrative flow data, supporting subsidy design, facility siting, capacity allocation, and accessibility-aware public-service planning.

\end{document}